\documentclass[preprint,12pt]{elsarticle}

\usepackage{amssymb}
\usepackage{amsmath}
\usepackage{multirow}

\usepackage{xcolor}

\journal{Medical Image Analysis}

\begin{document}

\begin{frontmatter}



\title{Learning Dissection Trajectories from Expert Surgical Videos via Imitation Learning with Equivariant Diffusion}



\author[a]{Hongyu Wang} 
\author[a]{Yonghao Long}
\author[a]{Yueyao Chen}
\author[b]{Hon-Chi Yip}
\author[c]{Markus Scheppach}
\author[d]{Philip Wai-Yan Chiu}
\author[d,e,f]{Yeung Yam}
\author[f]{Helen Mei-Ling Meng}
\author[a]{Qi Dou \corref{cor1}}

\affiliation[a]{organization={Department of Computer Science and Engineering, The Chinese University of Hong Kong},
            city={Hong Kong},
            country={China}}

\affiliation[b]{organization={Department of Surgery, The Chinese University of Hong Kong},
            city={Hong Kong},
            country={China}}
\affiliation[c]{organization={Internal Medicine III - Gastroenterology, University Hospital of Augsburg},
            city={Augsburg},
            country={Germany}}
\affiliation[d]{organization={Multi-scale Medical Robotics Center and The Chinese University of Hong Kong},
            city={Hong Kong},
            country={China}}
\affiliation[e]{organization={Department of Mechanical and Automation Engineering, The Chinese University of Hong Kong},
            city={Hong Kong},
            country={China}}
\affiliation[f]{organization={Centre for Perceptual and Interactive Intelligence and The Chinese University of Hong Kong},
            city={Hong Kong},
            country={China}}
\cortext[cor1]{Corresponding author: Qi Dou (qidou@cuhk.edu.hk)}

\begin{abstract}
Endoscopic Submucosal Dissection (ESD) constitutes a firmly well-established technique within endoscopic resection for the elimination of epithelial lesions.
Dissection trajectory prediction in ESD videos has the potential to strengthen surgical skills training and simplify surgical skills training. However, this approach has been seldom explored in previous research. While imitation learning has proven effective in learning skills from expert demonstrations, it encounters difficulties in predicting uncertain future movements, learning geometric symmetries and generalizing to diverse surgical scenarios. 
This paper introduces imitation learning for the critical task of predicting dissection trajectories from expert video demonstrations. We propose a novel Implicit Diffusion Policy with Equivariant Representations for Imitation Learning (iDPOE) to address this variability. Our method implicitly models expert behaviors using a joint state-action distribution, capturing the inherent stochasticity of future dissection trajectories and enabling robust visual representation learning across various endoscopic views. By incorporating a diffusion model in policy learning, our approach facilitates efficient training and sampling, resulting in more accurate predictions and improved generalization. Additionally, we integrate equivariance into the learning process to enhance the model's ability to generalize to geometric symmetries in trajectory prediction. To enable conditional sampling from the implicit policy, we develop a forward-process guided action inference strategy to correct state mismatches. 
We evaluated our method using a collected ESD video dataset comprising nearly 2000 clips. Experimental results demonstrate that our approach outperforms both explicit and implicit state-of-the-art methods in trajectory prediction. As far as we know, this is the first endeavor to utilize imitation learning-based techniques for surgical skill learning in terms of dissection trajectory prediction.
\end{abstract}





\end{frontmatter}



\section{Introduction}
\label{sec1}


Although deep learning-based approaches have demonstrated significant potential in surgical scene analysis \citep{maier2017surgical, loftus2020artificial, maier2022surgical}, enhancing aspects such as intelligent workflow recognition \cite{garrow2021workflow, jin2022trans} and scene understanding \citep{allan2020scenesegmentation, nwoye2022rendezvous}, research on advanced precise assitance for surgical procedures remains limited. One of the most critical tasks involves aiding decision-making regarding dissection trajectories \cite{wang2022real, guo2020novel, qin2020davincinet}, which is essential for ensuring surgical safety. 
Endoscopic Submucosal Dissection (ESD), a procedure used to treat early stage gastrointestinal cancers \cite{zhang2020symmetric, lau2021advanced}, involves multiple dissection maneuvers that demand substantial experience to identify the optimal path and impose significant stress on surgeons. 
Providing informative suggestions on dissection trajectories can greatly assist surgeons by reducing operative errors \cite{kim2011factors}, minimizing the risk of complications, and providing feedback on surgical skills training \cite{laurence2012laparoscopic}.
However, predicting the optimal path based on endoscopic video is complex. 
Firstly, determining dissection trajectories is intricate and challenging for even surgeon experts with many years’ experience due to numerous factors, such as the safety margins around the tumor. Secondly, blurred scenes and poor visual conditions can further impede scene recognition \cite{wang2022landmark}. 
To date, no data-driven solutions have been developed to predict dissection trajectories. We argue that it is feasible to learn this skill from expert video demonstrations.

Imitation learning has been extensively researched across various fields due to its strong ability to acquire complex skills \cite{hussein2017imitation, klaser2021imitation_medical, le2022survey_driving}. However, it requires adaptation and enhancement when applied to learn dissection trajectories from surgical data. One major challenge is the inherent uncertainty of future trajectories. Supervised methods, such as Behavior Cloning (BC) \cite{codevilla2019exploring}, tend to average all potential prediction paths, resulting in inaccurate forecasts. Although advanced probabilistic models aim to capture the complexity and variability of dissection trajectories \cite{li2017infogail, ren2021generalization, ke2021imitation}, ensuring reliable predictions across different surgical scenarios remains a significant challenge. To address these issues, implicit models are being developed to represent the policy, leading us to adopt Implicit Behavior Cloning (iBC) \cite{florence2022ibc}. iBC can learn robust representations by capturing the shared features of visual inputs and trajectory predictions through a unified implicit function, providing superior expressivity and improved visual generalization. Nonetheless, these methods have limitations. For example, techniques utilizing energy-based models (EBMs) \cite{florence2022ibc, jarrett2020strictly, ganapathi2022implicit, du2019implicit} require intensive computations due to their reliance on Langevin dynamics, causing a slow training process. Moreover, their performance can be sensitive to data distribution, and noise in the training data can lead to unstable prediction results of trajectories.
In addition, since trajectory prediction tasks inherently encompass geometric symmetries such as rotations, learning implicit policies also has the limitation that their optimization is more complex than learning explicit policies, making it more difficult to leverage symmetries underlying the task.

\begin{figure*}[t!]
    \centering
    \includegraphics[width=\textwidth]{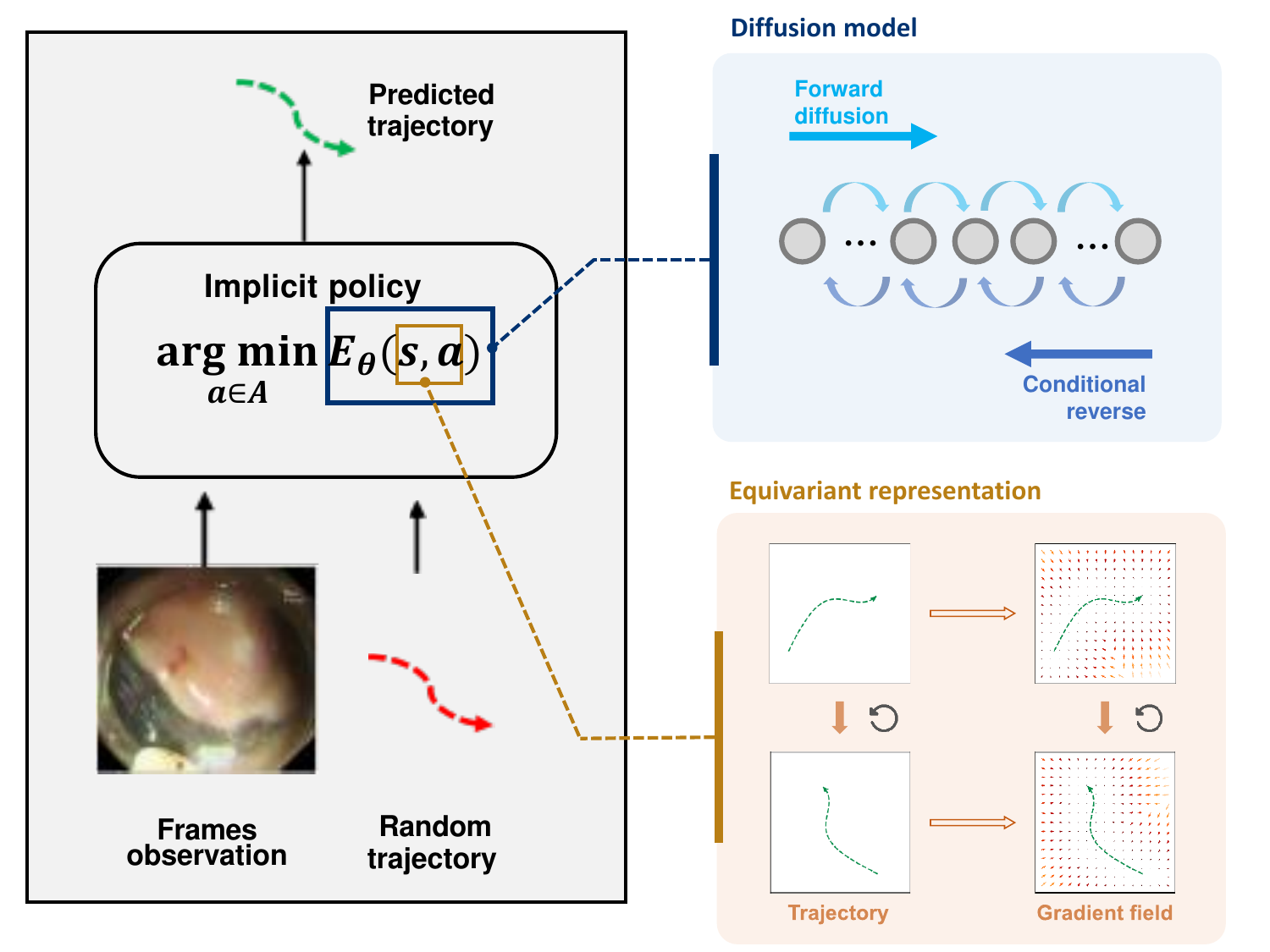}
    \caption{A graphical abstract of iDPOE.  iDPOE predicts dissection trajectories by learning the implicit policy as a diffusion model with equivariant representations.}
    \label{Fig1}
\end{figure*}

In this paper, we investigate the task of predicting dissection trajectories in endoscopic submucosal dissection surgery using imitation learning on expert video data. We present a novel method called Implicit Diffusion Policy with Equivariant Representations for Imitation Learning (iDPOE) for this purpose. The graphical abstract of our method is depicted in Fig.\ref{Fig1}. Firstly, to effectively model the surgeon's behaviors and learn the significant variation in surgical scenes, we employ implicit modeling to represent expert dissection skills. Secondly, to overcome the inefficient training and unstable performance issues associated with implicit policies by energy-based models, we formulate the implicit policy using an unconditional diffusion model, which excels in representing complex high-dimensional data distributions such as images or videos. Additionally, we integrate rotational symmetry into the diffusion model, allowing it to learn the equivariance properties of dissection trajectories. Furthermore, we develop a conditional action inference strategy guided by forward-diffusion to generate predictions from the implicit policy. To evaluate the effectiveness of our method, we curated a surgical video dataset of ESD procedures, comprising nearly two thousand annotated dissection trajectories. Our results demonstrate that our method outperforms state-of-the-art trajectory prediction methods across various surgical scenarios.
Our main contributions are as follows: (1) we propose to use diffusion models as a powerful implicit policy learning method for surgical trajectory prediction. Our diffusion-based method enables efficient modeling of complex surgical trajectories directly from high-dimensional endoscopic videos. (2) we propose to explicitly embed geometric equivariance into a diffusion-based imitation learning framework. By explicitly modeling geometric symmetries inherent in dissection trajectories, our method enhances trajectory prediction performance across varied surgical contexts. (3) we evaluate our method comprehensively on real-world endoscopic surgical video datasets. Our experimental results clearly demonstrate the superior performance of our method in trajectory prediction, generalization ability, and robustness compared to prior methods.

A preliminary version of this work was presented in MICCAI 2023 \cite{li2023imitation}. 
To further advance our study, we have substantially revised and extend the conference paper. This paper introduces the following extensions to our previous work: (1) we further improve our method by incorporating equivariance in the reverse process of diffusion model for policy learning; (2) We add more ESD surgery cases and conduct comprehensive experiments to evaluate the effectiveness of our method on the extended dataset; (3) our method outperforms competing methods on the dissection trajectory prediction task; and (4) we discuss our method in more details.

\section{Related Work}
\label{sec2}

\subsection{Intelligent Guidance for Endoscopic Submucosal Dissection}
Deep learning-based methods have achieved encouraging performance for many applications of automated medical image interpretation, and thus provide promising solutions for guidance in ESD surgery scenarios. Zhu et al. \cite{ZHU2019806} presented a method consisting of a convolutional neural network computer-aided detection system based on endoscopic images to accurately determine the invasion depth of gastric cancer. Xia et al. \cite{xia2021use} developed an AI-based auxiliary diagnostic system for magnetically controlled capsule endoscopy to detect gastric lesions, which showed good performance in sensitivity, specificity, and accuracy. Yang et al. \cite{yang2023uncertainty} proposed a novel fine-grained reward maximization framework for uncertainty estimation in safety-critical scene segmentation, which utilizes an uncertainty metric related reward function and a fine-grained parameter update scheme. Ghatwary et al. \cite{ghatwary2019early} evaluated the performance of different deep learning object detection models to automatically detect esophageal adenocarcinoma abnormalities from high-definition white light endoscopy images. Cao et al. \cite{cao2023intelligent} proposed an intelligent surgical workflow recognition suit for endoscopic submucosal dissection and integrated into real-time animal studies with promising results. These studies show promising results in various applications for ESD. By contrast, we focus on facilitating precise dissection trajectory prediction using deep learning-based frameworks.
The methods proposed in studies \cite{ZHU2019806, xia2021use} are designed for preoperative assessment, primarily focusing on classification and detection tasks, which serve as a foundation for intraoperative guidance. In contrast, studies \cite{yang2023uncertainty, ghatwary2019early, cao2023intelligent} address intraoperative guidance but are limited to segmentation or phase recognition. Our work also targets intraoperative guidance; however, it specifically aims to enhance precise dissection trajectory prediction.

\subsection{Imitation learning}
Imitation learning has shown a promising pathway to learning complicated skills from expert demonstrations. The imitation learning-based methods can mainly be grouped into two categories according to their form of learning policy. A range of studies, named explicit policy, attempt to learn action representations directly from observation states. Toyer et al. \cite{toyer2020magical} introduced the MAGICAL benchmark for evaluating the robustness of Imitation Learning algorithms, showing that existing algorithms fail to generalize beyond the demonstration variant and that common modifications have limited success in improving generalization. Shafiullah et al. \cite{shafiullah2022behavior} presented Behavior Transformers, a new technique to model unlabeled demonstration data with multiple modes for behavior learning, which outperforms prior methods and captures the major modes in the data, and analyzes the importance of its components through ablation studies. Zeng et al. \cite{zeng2021transporter} presented the Transporter Network for vision-based robotic manipulation that infers spatial displacements from visual input, showing high sample efficiency and generalization in both simulation and real-world experiments. The family of implicit policy-based approaches aims to define action distributions by learning an energy-based function.  Du et al. \cite{du2021improved} proposed an improved contrastive divergence training framework for Energy-Based Models by estimating a neglected gradient term, using data augmentation and multi-scale processing.
Mandlekar et al. \cite{mandlekar2022matters} conducted an extensive study of offline learning algorithms for robot manipulation on various tasks and datasets, and derives lessons including the effectiveness of history-dependent models, the importance of observation space and hyperparameters, and the potential of large human datasets for solving complex tasks.
Existing imitation learning methods suffer from computationally expensive training, sensitivity to data variations, and inability to explicitly utilize geometric symmetries, which limits their scalability and generalization in surgical scenarios. To address these limitations, we propose leveraging diffusion models for efficient training and  explicitly incorporating equivariance to generalize across various surgical contexts.

\subsection{Equivariant Deep Learning}
The movements of instruments within a two-dimensional Euclidean space inherently involve symmetries such as planar rotations. To enable models to preserve equivariance for these geometric symmetries is an important task in trajectory prediction.  Cohen et al. \cite{cohen2016group} extended traditional convolutional neural networks by incorporating symmetries such as rotations and reflections, which enhances learning efficiency without increasing the parameter count.  Finzi et al. \cite{finzi2020generalizing} proposed a convolutional layer equivariant to Lie groups for arbitrary continuous data, and demonstrates its performance on image, molecular, and dynamical system tasks.
Fuchs \cite{fuchs2020se} introduced the SE(3)-Transformer, a self-attention mechanism for 3D point clouds and graphs that is equivariant under 3D rotations. Dehmamy \cite{dehmamy2021automatic} proposed the Lie algebra convolutional network as a building block for constructing group equivariant neural networks, which can automatically discover symmetries, approximate group convolutions, and has connections to physics. Ouderaa et al. \cite{van2024learning} proposed a method to automatically learn layer-wise equivariance in neural networks using gradients, by improving relaxed equivariance and learning the amount of equivariance through Bayesian model selection.  Wang et al. \cite{wang2022approximately} defined approximately equivariant networks by relaxing equivariance constraints in group convolution, and demonstrates their superiority in modeling imperfectly symmetric dynamics in simulated and real-world datasets.
Despite the success of equivariant methods in various domains, their potential for modeling geometric symmetries inherent in endoscopic procedures remains underexplored, leading to limited generalization in existing surgical imitation learning approaches. To address this, we explicitly embed rotational symmetry into the learning of implicit diffusion policy, enabling the model to effectively capture geometric symmetries.

\subsection{Diffusion Model}
Diffusion models are powerful generative models that learn the reverse processes by refining randomly sampled Gaussian noises. On the other hand, diffusion models can be viewed as a process of implicit action scores and show their capacity to learn policies. Xu et al. \cite{xugeodiff2022} proposed GEODIFF, a novel generative model for molecular conformation prediction based on denoising diffusion models directly acting on atomic coordinates. Wang et al. \cite{wangdiffusion2023} proposed Diffusion Q-learning (Diffusion-QL), an offline reinforcement learning algorithm using a conditional diffusion model for policy regularization and Q-learning guidance. Ajay et al. \cite{ajayconditional2023} proposed the Decision Diffuser, a conditional generative model for sequential decision-making that frames offline decision-making as conditional generative modeling. Janner et al. \cite{janner2022planning} proposed a trajectory-level diffusion probabilistic model for data-driven trajectory optimization. Xian et al. \cite{xian2023chaineddiffuser} presented a neural policy architecture that unifies macro-action prediction and conditional trajectory diffusion for learning robot manipulation from demonstrations.
While diffusion models have shown strong capabilities in modeling complex trajectory distributions, their direct application to imitation learning for surgical applications is challenged by scene variability, geometric symmetries, and the high-dimensional nature of endoscopic surgical videos. To overcome these limitations, we aim to leverage implicit diffusion policies integrated with equivariant representations, enabling generalization to geometric symmetries, and accurate trajectory prediction in surgical scenarios.

\section{Method}
In this section, we introduce our approach, iDPOE, which uses an implicit diffusion policy to learn dissection trajectory prediction from expert video data. First, we present the formulation of the implicit policy for imitation learning of dissection trajectories. Next, we explain the training process for the implicit policy as an unconditional generative diffusion model. Finally, we describe the action inference strategy guided by forward-diffusion, enabling accurate trajectory prediction with the implicit diffusion policy.

\begin{figure*}[t!]
    \centering
    \includegraphics[width=\textwidth]{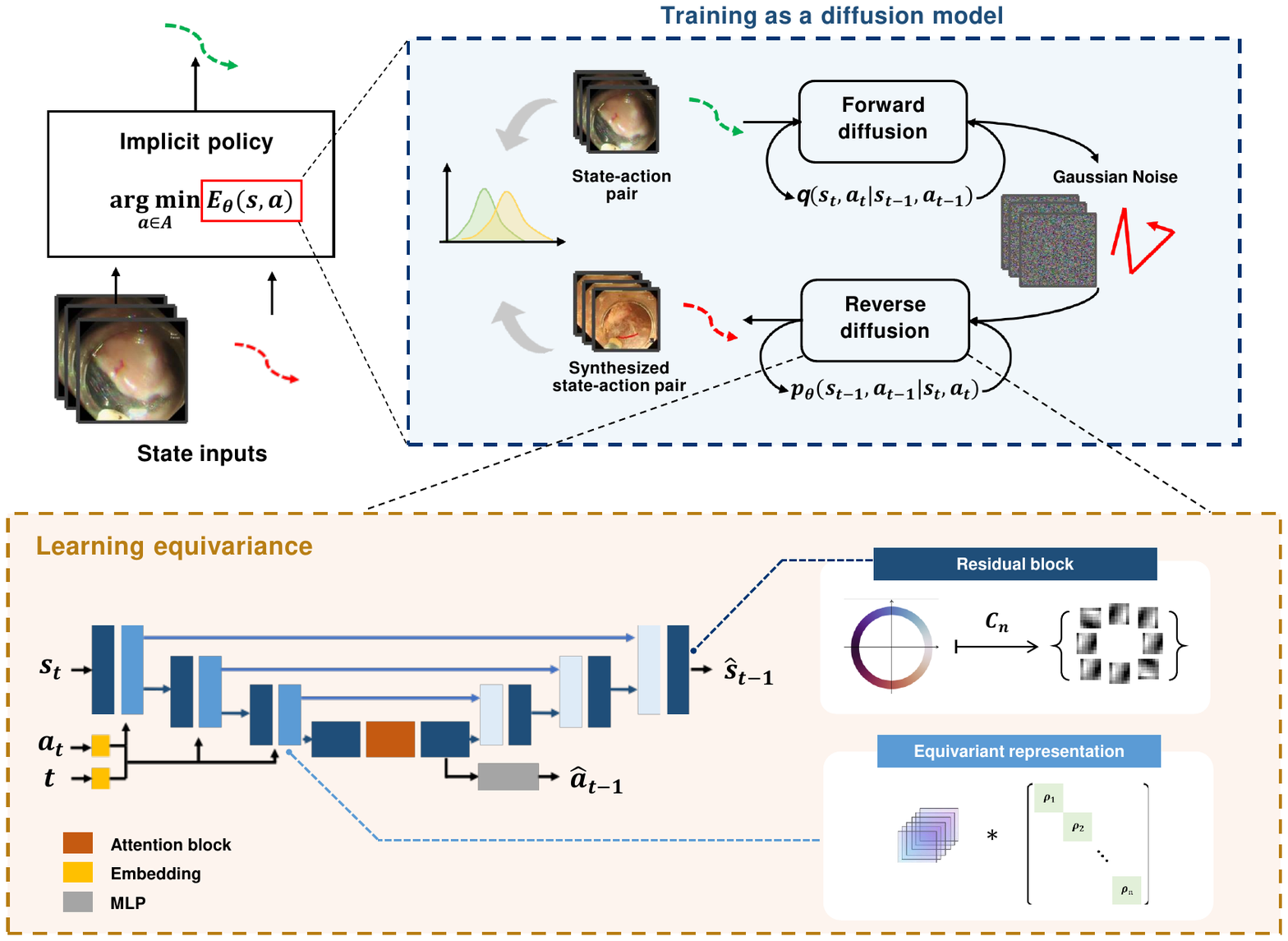}
    \caption{An overview of our approach iDPOE. (a) illustrates the modeling of the implicit policy, and its training process. We train a diffusion model with equivariant representations to approximate the joint state-action distribution, and (b) depicts the inference loop for trajectory prediction with the forward-diffusion guidance.}
    \label{Fig2}
\end{figure*}

\subsection{Problem Definition}
Our approach formulates the prediction of dissection trajectories as an imitation learning problem based on expert demonstrations. We define a Markov Decision Process (MDP) $\mathcal{M}=(\mathcal{S}, \mathcal{A}, \mathcal{T}, \mathcal{D})$, which includes the state space $\mathcal{S}$, action set $\mathcal{A}$, state transition distribution $\mathcal{T}$, and expert demonstrations $\mathcal{D}$. The objective of our method is to learn a prediction policy $\pi^*(a|s)$ from the expert demonstrations $\mathcal{D}$. The input state for the policy consists of a clip of video frames $s=\{I_{t-L+1}, I_{t-L+2}, \dots, I_{t}\}$, where each frame $I_t \in \mathbb{R}^{H\times W\times 3}$ and $L$ represents the length of the video. The output is a sequence of action coordinates $a = \{y_{t+1}, y_{t+2}, ..., y_{t+N}\}$, where each $y_t \in \mathbb{R}^2$ and $N$ denotes the number of action steps, indicating the future dissection trajectory projected onto the image space.

To obtain the demonstrated dissection trajectories from expert video data, we manually annotate the dissection trajectories on video data based on the movement patterns of the instruments observed in subsequent frames. These annotations are used to construct a dataset $\mathcal{D}=\{(s, a)_i\}_{i=0}^M$ comprising $M$ pairs of video clips (states) and dissection trajectories (actions). To accurately model the expert's dissection behavior and effectively learn generalizable features from the demonstrations, we employ implicit modeling as our imitation policy. Building on the formulation presented in \cite{florence2022ibc}, we define the dissection trajectory prediction policy as the maximization of the joint state-action density function $\arg\max_{a\in \mathcal{A}} p_\theta(s, a)$ rather than an explicit function $F_\theta(s)$, as illustrated in Fig. \ref{Fig2}. Here, $\hat{a}$ represents the optimal action derived from the policy distribution conditioned on the state $s$, and $p_\theta(s, a)$ denotes the joint state-action distribution.

We employ the behavior cloning objective to learn the implicit policy from expert demonstrations by minimizing the Kullback-Leibler (KL) divergence between the learned policy $\pi_\theta(a|s)$ and the distribution of the demonstrations $\mathcal{D}$. This objective is also equivalent to maximizing the expected log-likelihood of the joint state-action distribution, as shown:
\begin{equation}\label{eq.imitation_learning_objective}
    \max_\theta \mathbb{E}_{(s,a) \sim \mathcal{D}} [\log \pi_\theta(a|s)]=\max_\theta \mathbb{E}_{(s,a) \sim \mathcal{D}} [\log p_\theta(s, a)].
\end{equation}
Hence, imitating dissection decision-making of surgeons is transformed into a distribution approximation problem.

\subsection{Learning Implicit Policy as Diffusion Models}
\label{sec3.2}
Approximating the joint state-action distribution in Eq.~\ref{eq.imitation_learning_objective} from video demonstration data has been challenging for previous EBM-based methods. To address this, we leverage recent advances in diffusion models to learn the implicit policy. These models represent the data through a continuous thermodynamics diffusion process, which can be discretized into a series of Gaussian transitions. This approach allows the diffusion model to express complex high-dimensional distributions with simple parameterized functions. Additionally, the diffusion process acts as a form of data augmentation by adding varying levels of noise to the data, enhancing generalization in high-dimensional state space.

As depicted in Fig.~\ref{Fig2}, the diffusion model includes a predefined forward diffusion process and a learnable reverse denoising process. The forward process incrementally diffuses the original data $x_0=(s,a)$ into a series of noisy data $\{x_0,x_1,\cdots,x_T\}$ using a Gaussian kernel $q(x_t|x_{t-1})$,  where T represents the diffusion steps. During the reverse process, the data is iteratively recovered using a parameterized Gaussian $p_\theta(x_{t-1}|x_t)$. Note that the reverse process is a trainable Gaussian distribution whose mean and variance are parameterized by a neural network.


To train the implicit diffusion policy, we aim to maximize the log-likelihood of the state-action distribution in Eq.\ref{eq.imitation_learning_objective}. By employing the Evidence Lower Bound (ELBO) as a proxy, the likelihood maximization is simplified to a noise prediction problem, as detailed in \cite{ho2020DDPM}. The noise prediction errors for the state and action are combined using a weight $\gamma \in [0, 1]$ as follows:
\begin{equation}
    \mathcal{L}_{noise}(\theta) = \mathrm{E}_{\epsilon, t, x_0}[(1 - \gamma)\|\epsilon^a_\theta(x_t, t) - \epsilon^a\| + \gamma\|\epsilon^s_\theta(x_t, t) - \epsilon^s\|],
\end{equation}
where $\epsilon^s$ and $\epsilon^a$ are sampled from $\mathcal{N}(0,\textit{\textbf{I}}^s)$, $\mathcal{N}(0, \textit{\textbf{I}}^a)$ respectively. To more effectively process features from video frames and coordinate trajectories, we employ a U-Net-based model as the implicit diffusion policy network. The trajectory information is integrated into the feature channels through MLP embedding layers, and the trajectory noise is predicted by an MLP branch at the bottleneck.

\subsection{Equivariant Learning on State Space}

An important step is to incorporate equivariance into the diffusion models for policy learning in order to learn the equivariant representations of state-action pairs $x=(s,a)$. A function f can be called G-equivariant if $f(g \cdot x)=g \cdot f(x)$, given $\forall g \in G$. Since equivariant neural networks are able to learn to maintain equivariance with respect to rotational symmetries \cite{e2cnn2019}, we aim to allow the learned policies to generalize across the rotational transformations of inputs by incorporating equivariance.

Since equivariant neural networks aim to describe group-equivariant operations on the image plane $\mathbb{R}^2$, their feature spaces are typically defined as the spaces of feature fields \cite{cesa2022a}. These feature fields can be characterized by group representations:  $\rho: G \rightarrow GL(\mathbb{R}^{|G|})$, where $|G|$ is the cardinality of $G$. Subsequently, to consider a group $G \subseteq SO(2)$ on the image plane, we first formalize the features on each position as feature vector fields, i.e. $f: \mathbb{R}^2 \rightarrow \mathbb{R}^c$, where $f$ maps each position of 2D space to a $c$-dimensional feature vector. Given that the group $G$ is a subgroup of $SO(2)$, a cyclic subgroup $C_n$ can be used as a specific instance of $G$ to describe $n$ discrete rotations because it is compatible with the symmetries of the grid properties of images and is easily implemented.
The key to constructing an equivariant neural network is that its formation of functions (i.e. each layer) is equivariant. The convolutional functions plays a role in adaptively exploring symmetries. More specifically, a convolutional operation with kernel $K$ is a mapping from the feature field $\mathbb{R}^{c_{in}}$ to the feature field $\mathbb{R}^{c_{out}}$. A convolutional operation is G-equivariant if and only if it satisfies the following constraint \cite{cohen2019general}:
\begin{equation}
     K(gx) = \rho_{out}(g) K(x) \rho_{in}(g^{-1}), \forall g \in G
\end{equation}
where $\rho_{in}(g)$ and $\rho_{out}(g)$ are the group representations of group elements on the feature fields $\mathbb{R}^{c_{in}}$ and $\mathbb{R}^{c_{out}}$, respectively. Since the feature spaces of neural networks are typically in the form of stacked independent features, it similarly requires that multiple feature fields are used to model the symmetric groups of transformations independently in the hidden layers of equivariant neural networks. The stacked feature fields are thus viewed as the direct sum of the individual group representations $\rho = \oplus_i \rho_i$. It is also proved that both the linear layers (e.g. fully-connected layers) and activation functions (e.g. ReLU function) are equivariant \cite{wang2021incorporating}. Recall that we employ a UNet-based model as the network for policy learning; therefore, the implicit diffusion policy network composed of such equivariant functions is adequate to learn equivariant representations.

\begin{figure*}[t!]
    \centering
    \includegraphics[width=\textwidth]{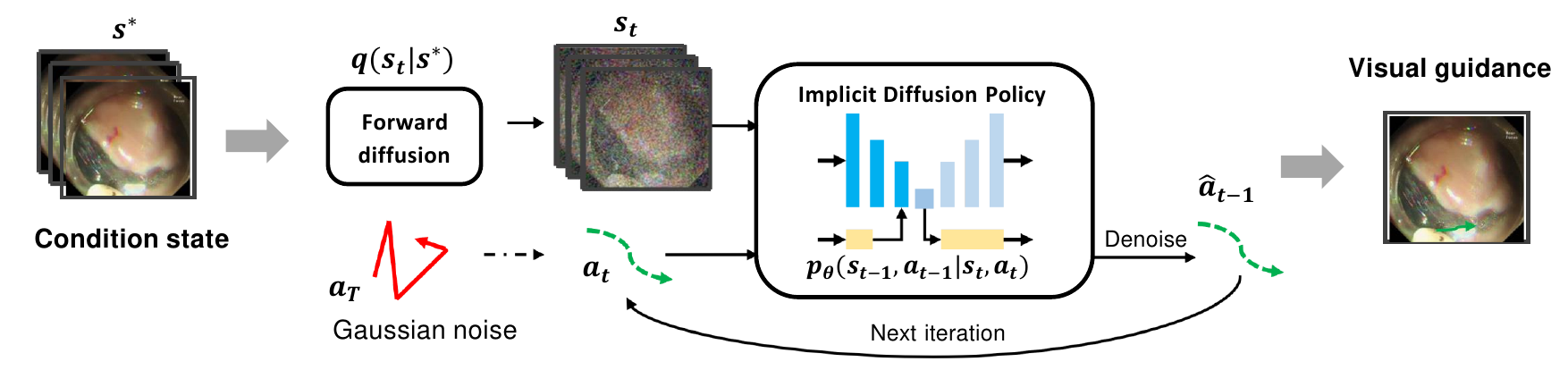}
    \caption{The inference process for trajectory prediction with the forward-diffusion guidance by conditional sampling.}
    \label{Fig2b}
\end{figure*}

\subsection{Conditional Sampling with Forward-Diffusion Guidance}
Since the training process described in Section~\ref{sec3.2} is designed for unconditional generation, the conventional sampling strategy through the reverse process will tend to predict random trajectories from the expert data. A straightforward method to introduce conditioning into the inference is to directly input the video clip as the condition state $s^*$ into the implicit diffusion policy and then sample only the action part. However, there is a distribution mismatch between the state $s^*$ and the $s_t$ in the training process, resulting in inaccurate predictions. Therefore, we propose a sampling strategy that corrects this distribution mismatch by incorporating forward-process guidance into the reverse sampling procedure.

Considering the reverse process of the diffusion model, the transition probability conditioned on $s^*$ can be decomposed into
\begin{equation}
     p_\theta(x_{t-1}|x_t, s^*) = p_\theta(x_{t-1}|s_t, a_t, s^*) = p_\theta(x_{t-1}|x_t)q(s_t|s^*),
\end{equation}
where $x_t = (s_t, a_t)$, $p_\theta(x_{t-1}|x_t)$ denotes the learned denoising function of the implicit diffusion model, and $q(s_t|s^*)$ represents a forward diffusion process from the conditioned state to the $t$-th diffused state. Therefore, we can achieve conditional sampling by incorporating forward-process guidance into the reverse sampling process of the diffusion model.

Figure~\ref{Fig2b} presents a schematic illustration of our sampling approach. At the initial step $t=T$, the action $a_T$ is sampled from pure Gaussian noise, while the input state $s_T$ is diffused from the input video clip $s^*$ through a forward diffusion process. At the $t$-th step of the denoising loop, the action input $a_t$ is derived from the denoised action of the previous step, whereas the visual inputs $s_t$ are still obtained from $s^*$ via the forward diffusion process. This forward diffusion process and the denoising step are repeated until $t=0$, at which point the final action $\hat{a}_0$ is the prediction from the implicit diffusion policy. A deterministic action can be obtained by taking the most probable samples during the reverse process.

\subsection{Implementation Details}
All experiments are performed using the PyTorch library. We use a variant of U-Net \cite{ronneberger2015u} with residual blocks and attention blocks as the network architecture of our method. For simplicity and efficiency, we implement the equivariant functions of the neural network under cyclic group $C_4$. In the training phase, we resize each image to $128 \times 128$, and then normalize its pixel values to the normal distribution. We use the Adam optimizer with default hyper-parameters of $\beta_1 = 0.9$ , $\beta_2 = 0.999$. We set the batch size to 32, learning rate to 0.0001, and maximum epoch to 200. The learning rate is decayed via a cosine annealing schedule. The validation set is used to monitor the model training. In the test phase, each test image is also resized to $128 \times 128$ and three consecutive frames are fed into the trained model to predict the trajectory in the future frames.

\section{Experiments}

\subsection{Dataset}
We evaluated the proposed approach using a dataset of 40 ESD surgery videos provided by the Endoscopy Centre of the Prince of Wales Hospital in Hong Kong. All videos were recorded using Olympus microscopes operated by an expert surgeon with 15 years of ESD experience. For inference speed, the original videos were downsampled to 2 frames per second (FPS). The input state is a 1.5-second video clip containing 3 consecutive frames, and the expert dissection trajectory is represented by a 6-point polyline indicating the tool's movements over the next 3 seconds. We annotated a total of 1993 video clips, each containing 3 frames. From 40 cases, 1351 clips (4053 frames) were selected for training, with 10\% used for validation, and 642 clips (1926 frames) were used for evaluation.

\begin{table*}[]
\label{tab1}
\centering
\caption{Quantitative results on the in-the-context and the out-of-the-context sets in metrics of ADE/FDE/FD. The lower is the better.}
\resizebox{\linewidth}{!}{
\renewcommand{\arraystretch}{1.5}{
\begin{tabular}{cccc|ccc}
\hline
                                          & \multicolumn{3}{c|}{In-the-context}                                                                                      & \multicolumn{3}{c}{Out-the-context}                                                                                      \\ \cline{2-7} 
\multirow{-2}{*}{Method}                  & ADE                                    & FDE                                    & FD                                     & ADE                                    & FDE                                    & FD                                     \\ \hline
BC                                        & 16.456 (±3.389)                        & 23.934 (±3.766)                        & 32.557 (±2.287)                        & 18.069 (±2.797)                        & 24.351 (±2.197)                        & 33.956 (±2.731)                        \\
iBC                                       & 17.960 (±3.291)                        & 26.878 (±5.171)                        & 37.537 (±4.091)                        & 21.678 (±3.445)                        & 27.962 (±5.216)                        & 36.134 (±4.101)                        \\
MID                                       & 18.372 (±3.318)                        & 25.332 (±3.773)                        & 35.613 (±2.118)                        & 20.103 (±1.892)                        & 26.177 (±2.983)                        & 35.595 (±4.157)                        \\
 LED              & 15.765 (±2.827) &  22.980 (±3.567) & 30.977 (±2.547) &  17.726 (±2.697) &  23.409 (±2.972) &  32.389 (±3.310) \\
 Singulartrajectory & 15.097 (±2.016) &  22.131 (±2.950) &  31.206 (±2.675) & 17.359 (±2.351) &  22.996 (±2.367) &  32.782 (±3.341) \\
Ours                                      & 13.340 (±1.543)                        & 20.211 (±2.501)                        & 29.397 (±2.119)                        & 15.576 (±1.557)                        & 22.076 (±1.080)                        & 31.182 (±2.235)                        \\ \hline
\end{tabular}
}
}
\end{table*}

\begin{figure*}[t]
    \centering
    \includegraphics[width=1.0\textwidth]{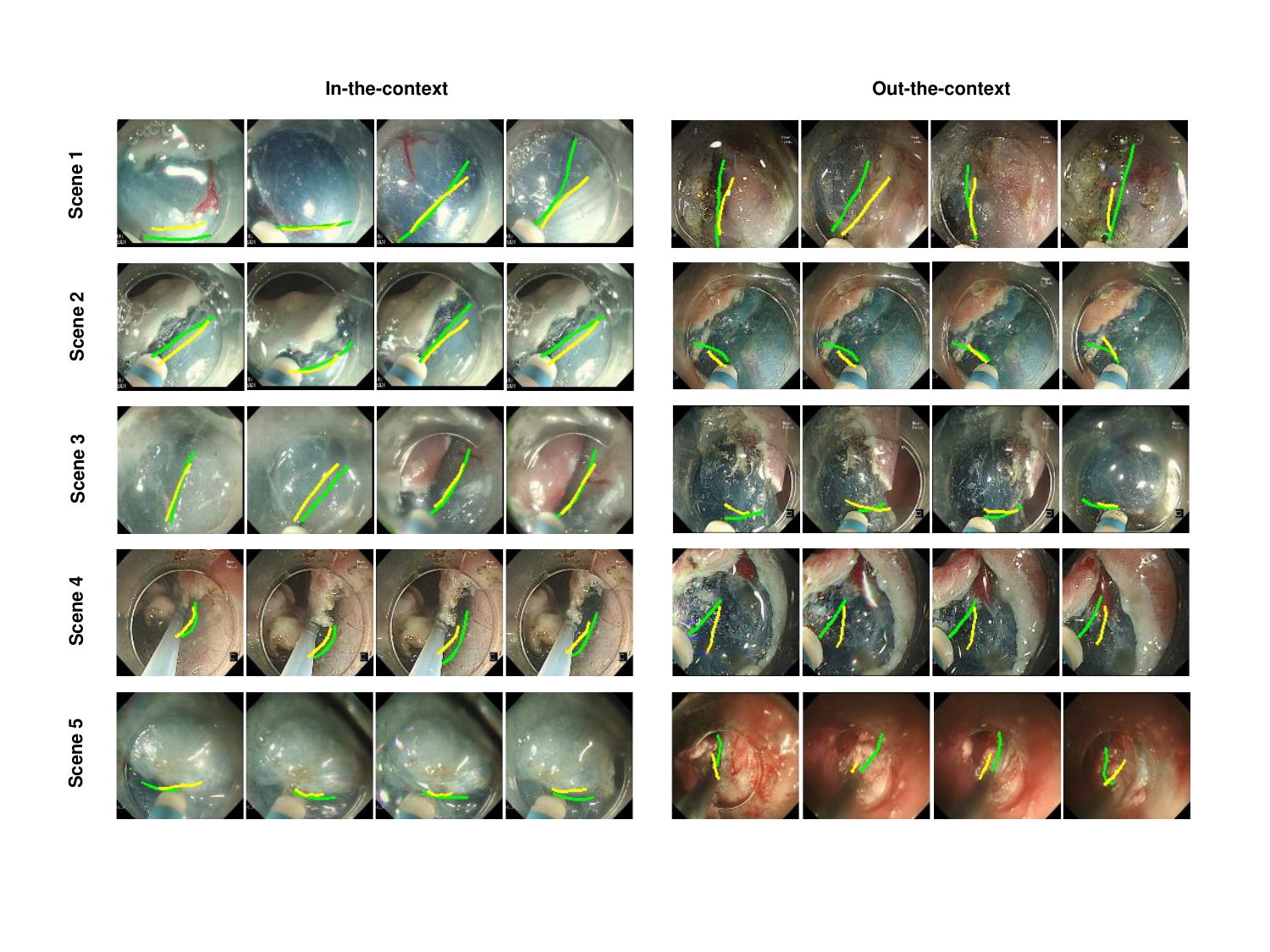}
    \caption{Visualized results of our method under settings of in-the-context and out-of-the-context evaluations. Green and yellow arrows respectively denote ground truths and predictions of dissection trajectory.}
    \label{Fig3}
\end{figure*}

\subsection{Experiment Settings}
First, to evaluate the model's performance on data within the same surgical context as the training data, we selected a subset referred to as the "in-the-context" set, which consists of consecutive frames selected from the same period in the training data. Second, to assess the model's ability to generalize to visually distinct scenes, we selected an "out-of-the-context" evaluation set composed of video clips from four previously unseen surgical cases, totaling 393 clips.

\subsection{Evaluation Metrics}
To assess the performance of the proposed approach, we employ several metrics, including commonly used evaluation measures for trajectory prediction \cite{mohamed2020social, sun2020recursive}. Average Displacement Error (ADE) reports the overall deviations between the predictions and the ground truth, while Final Displacement Error (FDE) describes the difference from the moving target by computing the L2 distance between the final trajectory points. Additionally, we utilize Fréchet Distance (FD) to indicate the geometrical similarity between two temporal sequences. Lower error values correspond to better performance. All the above metrics are measured in pixel errors.

\subsection{Comparison with State-of-the-Art Methods}
To evaluate the proposed approach, we selected a range of baselines and state-of-the-art methods for comparison. Behavior Cloning, a fully supervised method implemented using a CNN-MLP network, was chosen as the baseline. Additionally, we included iBC \cite{florence2022ibc}, an EBM-based implicit policy learning method, and MID \cite{gu2022mid}, a diffusion-based trajectory prediction approach, LED \cite{mao2023leapfrog}, a model that learn a diverse distribution in the reverse denoising process, and Singulartrajectory \cite{bae2024singulartrajectory}, a method that incorporates singular space into the diffusion process, as state-of-the-art comparison methods.

We compared the proposed method with state-of-the-art approaches both quantitatively and qualitatively. As shown in Table.~\ref{tab1}, our method outperforms these approaches in both "in-the-context" and "out-of-the-context" scenarios across all metrics. Compared to the diffusion-based method MID, iDPOE is more effective at predicting long-term goals, especially in "out-of-the-context" scenarios. The performance of iBC did not meet expectations, highlighting the limitations of EBM-based methods in learning visual representations from complex endoscopic scenes. 
LED fails to outperform due to the training instability caused by its Leapfrog initializer. The performance of SingularTrajectory indicates that the adaptive anchor points may not be adjusted effectively in dynamic scenarios, resulting in deviation from the predicted trajectory.
The superior results achieved by our method demonstrate the effectiveness of the diffusion model in learning the implicit policy from expert video data. Additionally, our method exhibits lower standard deviation in prediction errors, indicating its ability to learn generalizable dissection skills. The qualitative results are presented in Fig.~\ref{Fig3}. We selected three typical scenes in ESD surgery and displayed the predictions of iDPOE alongside the ground truth trajectories. The results show that our method can generate accurate visual guidance consistent with expert demonstrations in both evaluation sets.

\section{Discussion}
\subsection{Effect of Implicit Modeling}
\begin{figure*}[t]
    \centering
    \includegraphics[width=0.9\linewidth]{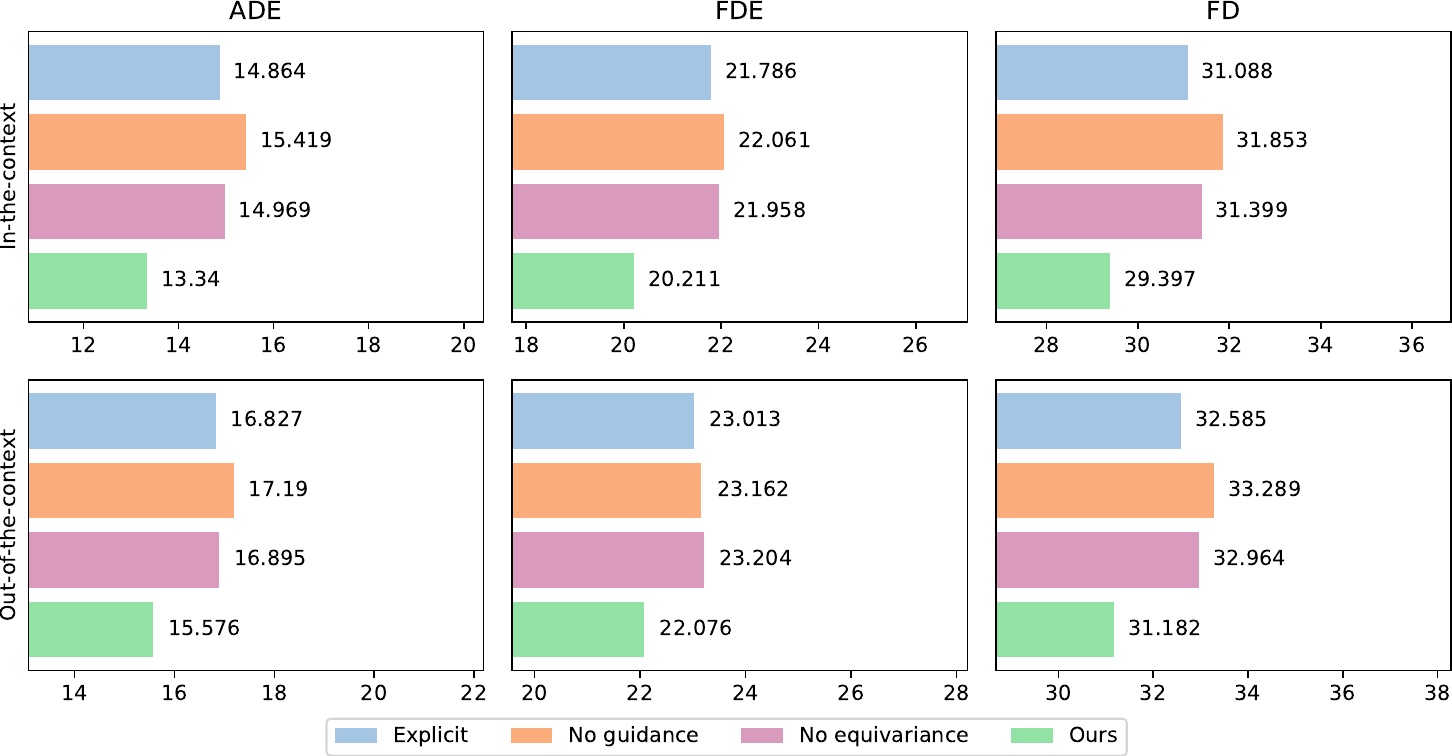}
    \caption{Ablation study for key components of our method.}
    \label{Fig_abl}
\end{figure*}

First, we investigated the significance of using implicit modeling for policy representation. We simulated the explicit form of the imitation policy by training a conditional diffusion model with a video clip as the conditional input. As shown in Fig.~\ref{Fig_abl}, the explicit diffusion policy performs worse on both evaluation sets compared to the implicit form. Implicit modeling significantly improves predictions in both "in-the-context" and "out-of-the-context" scenarios, indicating that the implicit model is better at capturing subtle changes in surgical scenes.

\subsection{Necessity of Forward-Diffusion Guidance}
\begin{figure*}[t]
    \centering
    \includegraphics[width=0.9\linewidth]{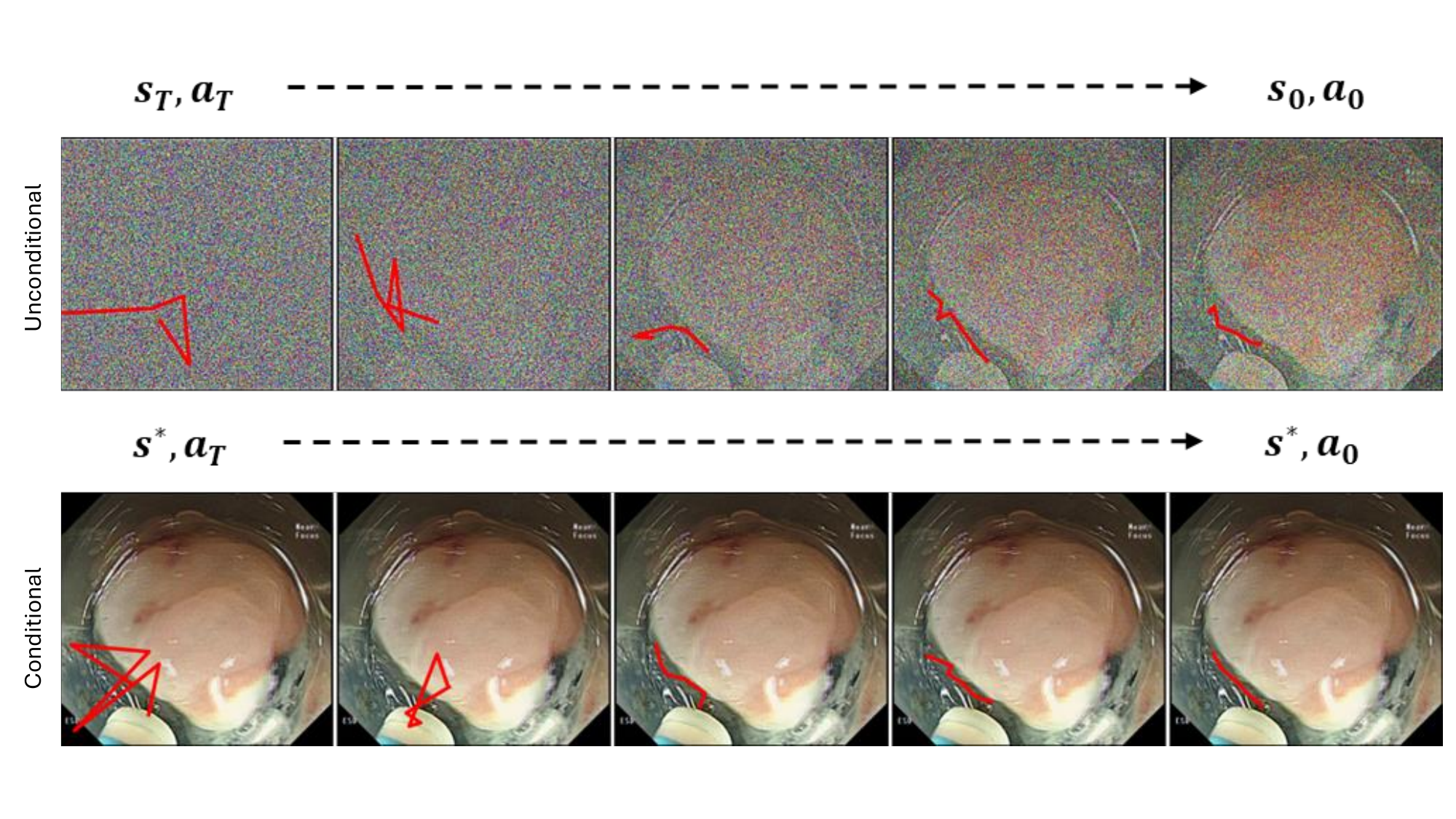}
    \caption{Visualization of reverse processes of unconditional/conditional sampling from implicit policy.}
    \label{Fig5}
\end{figure*}

We also explored the necessity of forward-diffusion guidance in conditional sampling for prediction accuracy and conducted an ablation study by predicting trajectories with unconditional sampling or with conditional sampling. By removing the forward-diffusion guidance during the action sampling procedure, we directly fed the condition state into the policy while sampling actions through the reverse process. As shown in Fig.~\ref{Fig_abl}, the implicit diffusion policy performs better with forward-diffusion guidance in both scenarios. 
For instance, without conditional sampling, the performance of our method drops by around 1.2-1.8 in the “out-of-the-context” scenario across all metrics. In addition, we also visualized the reverse process of unconditional and conditional sampling from implicit diffusion policy. As shown in Fig.~\ref{Fig5}, our method with conditional sampling can predict better trajectories compared with the one with unconditional sampling during the reverse process.
Thus, this suggests that the trajectory prediction inference strategy can be improved by incorporating forward-diffusion guidance.

\subsection{Effect of Equivariant representations}
Moreover, we analyzed the effectiveness of learning equivariant representations on performance. We compared our method to a variant of it without learning equivariant representations. As shown in Fig.~\ref{Fig_abl}, the implicit diffusion policy learning with equivariant representations performs better on both in-the-context and out-of-the-context sets. The results imply that incorporating equivariance into the implicit diffusion policy learning leads to better generalization.

\subsection{Value of Synthetic Data}
The learned implicit diffusion policy can naturally generate synthetic expert dissection trajectory data,  and these data are actually synthetic whose distributions actually differ from those of real-world data. These data could potentially reduce the high cost of manual annotation and help with training. To discuss the value of this synthetic data for downstream tasks, we trained the baseline model using the generated expert demonstrations. To validate how these generated data contribute to the training of our method, we randomly generated approximately 10,000 video-trajectory pairs through unconditional sampling from the implicit diffusion policy. We then trained the BC model with different datasets: pure expert data (real), synthetic data only (synt), and a mix of real and synthetic data (mix). The table in Table.~\ref{tab2} demonstrates that synthetic data is beneficial as augmented data for downstream task learning. 

\begin{table*}[]
\caption{Performance of BC trained with different data settings v.s. our method on ADE/FDE/FD.}
\label{tab2}
\resizebox{\linewidth}{!}{
\renewcommand{\arraystretch}{1.5}
\begin{tabular}{cccc|ccc}
\hline
\multirow{2}{*}{Method} & \multicolumn{3}{c|}{In-the-context}                 & \multicolumn{3}{c}{Out-the-context}                 \\ \cline{2-7} 
                        & ADE             & FDE             & FD              & ADE             & FDE             & FD              \\ \hline
BC (real)               & 16.456 (±3.389) & 23.934 (±3.766) & 32.557 (±2.287) & 18.069 (±2.797) & 24.351 (±2.197) & 33.956 (±2.731) \\
BC (synt)               & 18.523 (±2.873) & 24.863 (±3.701) & 34.459 (±3.663) & 19.496 (±3.072) & 25.817 (±3.338) & 35.223 (±3.107) \\
BC (mix)                & 15.654 (±2.873) & 22.637 (±2.237) & 31.552 (±2.331) & 17.790 (±2.519) & 23.650 (±2.002) & 33.256 (±2.752) \\
Ours                    & 13.340 (±1.543) & 20.211 (±2.501) & 29.397 (±2.119) & 15.576 (±1.557) & 22.076 (±1.080) & 31.182 (±2.235) \\ \hline
\end{tabular}
}
\end{table*}

\subsection{Effect of blurred visual conditions}
Due to fluid presence, rapid instrument movements, or suboptimal camera focus, blurred visual conditions may occur during endoscopic procedures. To discuss the robustness of our method under realistic blurred visual conditions, we conducted an additional experiment for robustness analysis.  Specifically, we selected a subset of blurred video frames from the original testing dataset, including 593 frames. We think that these blurred samples can ensure they accurately represent real-world scenarios encountered during ESD procedures. As shown in Table.~\ref{tab3}, all methods suffer degradation in performance under worse visual conditions. Moreover, our method suffers less performance degradation and still outperforms other methods, which demonstrates the better robustness of our method.

\begin{table*}[]
\caption{Quantitative results on the in-the-context and the out-of-the-context sets under blurred visual conditions.}
\label{tab3}
\resizebox{\linewidth}{!}{
\renewcommand{\arraystretch}{1.5}{
\begin{tabular}{cccc|ccc}
\hline
{\color[HTML]{333333} }                         & \multicolumn{3}{c|}{{\color[HTML]{333333} In-the-context}}                                                               & \multicolumn{3}{c}{{\color[HTML]{333333} Out-the-context}}                                                               \\ \cline{2-7} 
\multirow{-2}{*}{{\color[HTML]{333333} Method}} & {\color[HTML]{333333} ADE}             & {\color[HTML]{333333} FDE}             & {\color[HTML]{333333} FD}              & {\color[HTML]{333333} ADE}             & {\color[HTML]{333333} FDE}             & {\color[HTML]{333333} FD}              \\ \hline
{\color[HTML]{333333} BC}                       & {\color[HTML]{333333} 17.711 (±4.790)} & {\color[HTML]{333333} 24.973 (±4.925)} & {\color[HTML]{333333} 33.940 (±3.434)} & {\color[HTML]{333333} 19.383 (±4.099)} & {\color[HTML]{333333} 25.851 (±3.558)} & {\color[HTML]{333333} 34.962 (±3.741)} \\
{\color[HTML]{333333} iBC}                      & {\color[HTML]{333333} 19.262 (±4.348)} & {\color[HTML]{333333} 28.319 (±6.386)} & {\color[HTML]{333333} 38.676 (±5.248)} & {\color[HTML]{333333} 23.066 (±4.586)} & {\color[HTML]{333333} 29.413 (±6.635)} & {\color[HTML]{333333} 37.314 (±5.186)} \\
{\color[HTML]{333333} MID}                      & {\color[HTML]{333333} 19.574 (±4.326)} & {\color[HTML]{333333} 26.347 (±5.293)} & {\color[HTML]{333333} 37.043 (±3.414)} & {\color[HTML]{333333} 21.208 (±3.263)} & {\color[HTML]{333333} 27.278 (±4.301)} & {\color[HTML]{333333} 36.673 (±5.567)} \\
{\color[HTML]{333333} LED}                      & {\color[HTML]{333333} 17.029 (±4.115)} & {\color[HTML]{333333} 24.147 (±4.690)} & {\color[HTML]{333333} 32.385 (±3.567)} & {\color[HTML]{333333} 19.005 (±3.864)} & {\color[HTML]{333333} 24.731 (±4.009)} & {\color[HTML]{333333} 33.833 (±4.748)} \\
{\color[HTML]{333333} Singulartrajectory}       & {\color[HTML]{333333} 16.577 (±3.018)} & {\color[HTML]{333333} 23.616 (±4.177)} & {\color[HTML]{333333} 32.426 (±3.802)} & {\color[HTML]{333333} 18.605 (±3.511)} & {\color[HTML]{333333} 24.319 (±3.386)} & {\color[HTML]{333333} 34.098 (±4.414)} \\
{\color[HTML]{333333} Ours}                     & {\color[HTML]{333333} 14.049 (±2.894)} & {\color[HTML]{333333} 21.688 (±3.774)} & {\color[HTML]{333333} 30.692 (±3.211)} & {\color[HTML]{333333} 16.537 (±2.973)} & {\color[HTML]{333333} 23.561 (±2.226)} & {\color[HTML]{333333} 32.571 (±3.351)} \\ \hline
\end{tabular}
}}
\end{table*}

\subsection{Computational Complexity}
\begin{figure*}[]
    \centering
    \includegraphics[width=0.58\linewidth]{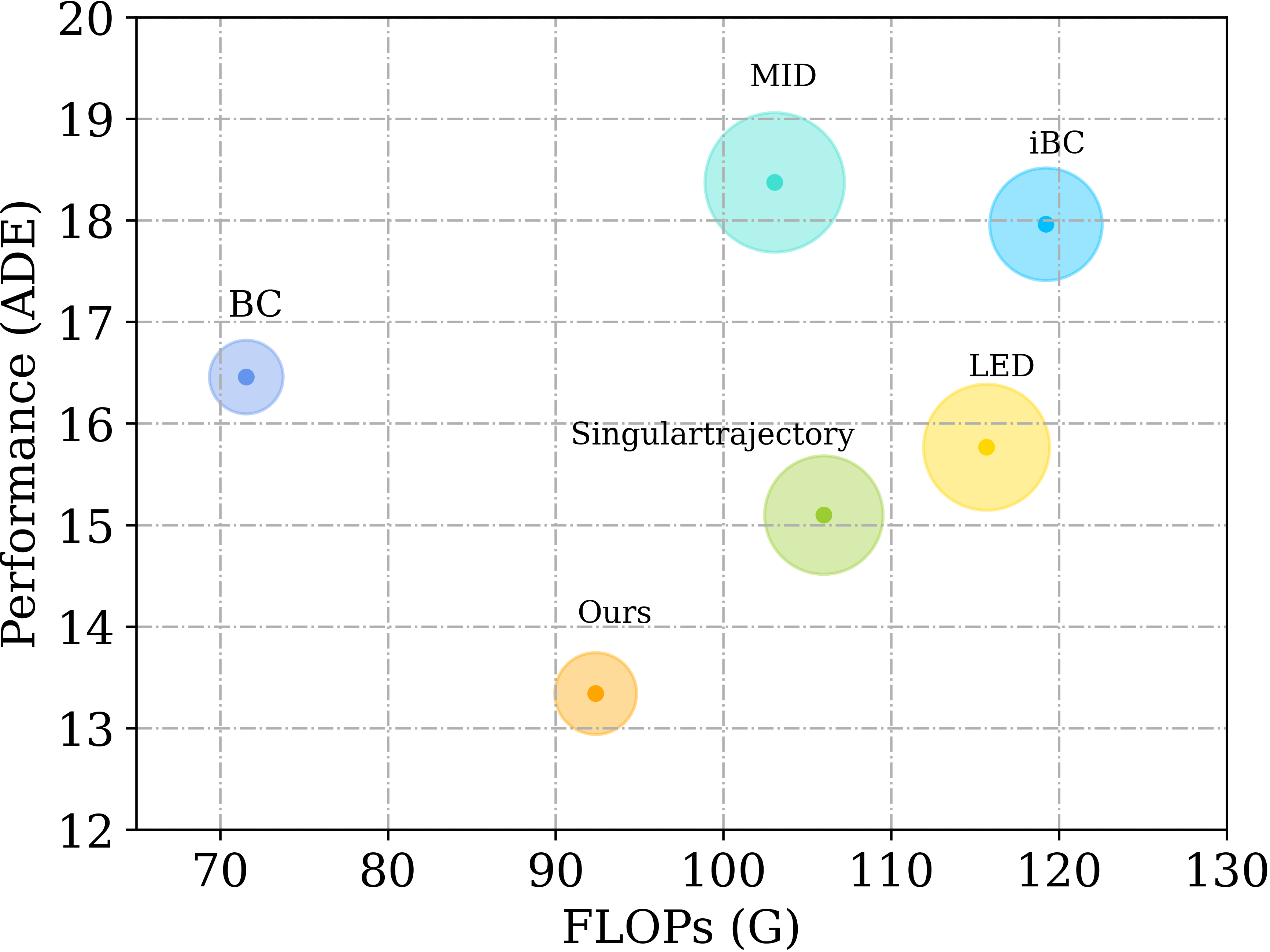}
    \caption{Performance versus computational complexity.}
    \label{Fig_comp}
\end{figure*}

We also discussed the computational complexity and inference speed of competing methods. The inference processes of all the competing methods were performed using three 128×128 video frames and under the same experimental environment (Intel Xeon E5-2678V32, NVIDIA 3090 RTX GPU, 128 GB Memory, and PyTorch library). We used floating-point operations per second (FLOPS) as a metric to show computational complexity. As shown in Fig. \ref{Fig_comp}, our method achieves better trade-off between computational complexity and prediction performance. In addition, our method surpasses other methods on the trajectory prediction task and achieves about 23 FPS inference speed compared to these competing methods while only slower than the BC model.

\section{Conclusion}
This paper introduces a novel approach, iDPOE, for learning dissection trajectory determination from expert video demonstrations. The proposed method leverages a diffusion model to represent the implicit policy, enabling the learning of equivariant properties of trajectories, which enhances the model's visual generalization capabilities. 
The iDPOE integrates the principles of implicit policy representation and diffusion modeling, which collectively contribute to its superior performance. By focusing on the equivariant properties of trajectories, the model not only learns the nuances of surgical dissection but also adapts to different visual contexts, making it robust against variations in surgical scenes. Our experimental results demonstrate that iDPOE outperforms state-of-the-art methods on the evaluation dataset, underscoring the effectiveness of our approach in learning dissection skills that generalize across various surgical scenarios. This work shows its potential for precise cognitive surgical assistance in real-world surgical applications.
In our future work, we will further validate the generalizability of iDPOE by extending evaluations to additional surgical procedures beyond ESD, such as laparoscopic colorectal resections. In addition, we aim to address the limitations regarding dataset accessibility by collaborating with our clinical research groups to potentially release a de-identified subset of our annotated data, subject to proper ethical approvals. To provide more valuable insights into the clinical applicability of our method, we also plan to benchmark the performance of iDPOE against surgeons with varying expertise levels (e.g., novices, senior, and expert surgeons).  We will actively seek collaborations with clinical surgeons to facilitate future comprehensive comparisons between our model and human performance. We hope to broaden the applicability of the iDPOE framework and facilitate its integration into diverse clinical settings.

\section*{Acknowledgement}
This research work was supported in part by the Hong Kong Innovation and Technology Fund (Project No. ITS/237/21FP), in part by the Research Grants Council of the Hong Kong Special Administrative Region, China (Project No. 24209223), in part by the National Natural Science Foundation of China (Project No. 62322318), and in part by the Dr. Stanley Ho Medical Development Foundation.







\bibliographystyle{elsarticle-num}
\bibliography{refs}



\end{document}